\documentclass{article}
\usepackage{ijcai22}

\usepackage{times}
\usepackage{soul}
\usepackage{url}
\usepackage[hidelinks]{hyperref}
\usepackage[utf8]{inputenc}
\usepackage[small]{caption}
\usepackage{graphicx}
\usepackage{amsmath}
\usepackage{amsthm}
\usepackage{booktabs}
\usepackage{algorithm}
\usepackage{algorithmic}

\usepackage{amssymb,mathrsfs}
\usepackage{color}
\usepackage{bbding}
\usepackage{multirow}
\usepackage{adjustbox}
\title{Lightweight Bimodal Network for Single-Image Super-Resolution via \\ Symmetric CNN and Recursive Transformer} 

\author{
Guangwei Gao$^{1,3}$\textsuperscript{$\dagger$}\and
Zhengxue Wang$^1$\textsuperscript{$\dagger$}\and
Juncheng Li$^2$\thanks{Corresponding author, $\dagger$Equal contribution}\and
Wenjie Li$^1$\and
Yi Yu$^3$\and
Tieyong Zeng$^2$
\affiliations
$^1$Nanjing University of Posts and Telecommunications\\
$^2$The Chinese University of Hong Kong $^3$National Institute of Informatics
\emails
\{csggao, cvjunchengli\}@gmail.com,
wzx\_0826@163.com
}

\begin{document}

\maketitle

\begin{abstract}
Single-image super-resolution (SISR) has achieved significant breakthroughs with the development of deep learning. However, these methods are difficult to be applied in real-world scenarios since they are inevitably accompanied by the problems of computational and memory costs caused by the complex operations. To solve this issue, we propose a Lightweight Bimodal Network (LBNet) for SISR. Specifically, an effective Symmetric CNN is designed for local feature extraction and coarse image reconstruction. Meanwhile, we propose a Recursive Transformer to fully learn the long-term dependence of images thus the global information can be fully used to further refine texture details. Studies show that the hybrid of CNN and Transformer can build a more efficient model. Extensive experiments have proved that our LBNet achieves more prominent performance than other state-of-the-art methods with a relatively low computational cost and memory consumption. The code is available at \url{https://github.com/IVIPLab/LBNet}.
\end{abstract}

\section{Introduction}
Single image super-resolution (SISR) aims to recover the corresponding high-resolution (HR) image with rich details and better visual quality from its degraded low-resolution (LR) one. Recently, convolutional neural networks (CNN) based SISR methods have achieved remarkable performance than traditional methods due to their powerful feature extraction ability. For example, Dong et al.~\cite{SRCNN} pioneered the Super-Resolution Convolutional Neural Network (SRCNN). Later, with the emergence of ResNet~\cite{ResNet} and DenseNet~\cite{DenseNet}, plenty of CNN-based SISR models have been proposed, like VDSR~\cite{VDSR}, EDSR~\cite{EDSR}, and RCAN~\cite{RCAN}. All these methods show that the deeper the network, the better the performance. However, these methods are difficult to be used in real-life scenarios with limited storage and computing capabilities. Therefore, a model that can achieve better performance while keeping the lightweight of the network has become attractive research. One of the most widely used strategies is to introduce the recursive mechanism, such as DRCN~\cite{DRCN} and DRRN~\cite{DRRN}. The other one is to explore the lightweight structure, including CARN~\cite{CARN}, FDIWN~\cite{gao2021feature}, and PFFN~\cite{zhang2021pffn}. Although these models reduce the number of model parameters to a certain extent through various strategies and structures, they also lead to a degradation in performance, thus it is difficult to reconstruct high-quality images with rich details.

\begin{figure*}[t]
	\centerline{\includegraphics[width=17cm, trim=0 20 0 0]{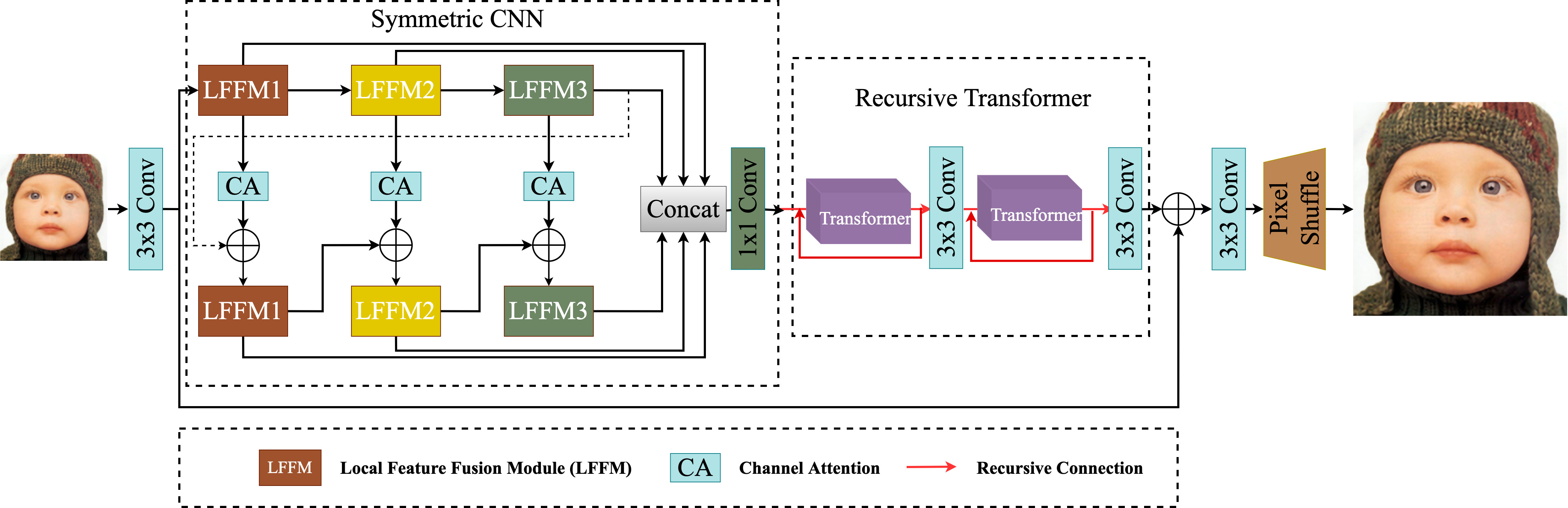}}
	\caption{The complete architecture of the proposed Lightweight Bimodal Network (LBNet).}
	\vspace{-0.3cm}
	\label{LBNet}
\end{figure*}

Recently, with the continuous progress of Transformer in Natural Language Processing (NLP), how to apply it to computer vision tasks has become a hot topic. Transformer can model the long-term dependence in the image, and this powerful representation ability can help to restore the texture details of the image. However, most methods blindly use Transformer to replace all original CNN structures, this is unreasonable since the ability of CNN to extract local features is irreplaceable. These features can maintain their own stability under different viewing angles, also called local invariance, which is helpful for image understanding and reconstruction. Therefore, we recommend fusing CNN and Transformer to take full use of the advantages of both to achieve efficient SR image reconstruction. 

To achieve this, we propose a Lightweight Bimodal Network (LBNet) for SISR. In LBNet, we use both CNN and Transformer to achieve dual-mode coordination reconstruction. As for the CNN part, we focus on local features extraction. Specifically, we propose a novel Local Feature Fusion Module (LFFM), which consists of a series of Feature Refinement Dual-Attention Block (FRDAB). FRDAB uses the channel reduction strategy to reduce the parameters of the model and introduces channel attention and spatial attention mechanisms to reweight the feature information extracted from different branches. Meanwhile, to balance the performance and size of the model, we introduce the parameter sharing strategy to construct the symmetric-like network, the output of the corresponding shared module of the previous stage will be integrated through the channel attention module as the input of the current module. This method can maximize the use of feed-forward features to restore texture details. As for the Transformer part, we propose a Recursive Transformer to learn the long-term dependence of images, thus the texture details can be further refined with global information. In summary, the main contributions are as follows
\begin{itemize}
  \item  We propose an effective Symmetric CNN for local feature extraction and coarse image reconstruction. Among them, Local Feature Fusion Module (LFFM) and Feature Refinement Dual-Attention Block (FRDAB) are specially designed for feature extraction and utilization.
  \item We propose a Recursive Transformer to learn the long-term dependence of images. This is the first attempt of the recursive mechanism in Transformer, which can refine the texture details by global information with few parameters and GPU memory consumption.
  \item We propose a novel Lightweight Bimodal Network (LBNet) for SISR. LBNet elegantly integrates CNN and Transformer, enabling it to achieve a better balance between the performance, size, execution time, and GPU memory consumption of the model.
\end{itemize}

\section{Related Works}
\subsection{CNN-based SISR}
Thanks to the powerful feature representation and learning capabilities of CNN, CNN-based SISR methods have made great progress in recent years~\cite{li2021beginner}. For example, SRCNN~\cite{SRCNN} applied CNN to SISR for the first time and achieved competitive performance at the time. EDSR~\cite{EDSR} greatly improved the model performance by using the residual blocks~\cite{ResNet}. RCAN~\cite{RCAN} introduced the channel attention mechanism and built an 800-layer network. Apart from these deep networks, many lightweight SISR models also have been proposed in recent years. For instance, Ahn et al.~\cite{CARN} proposed a lightweight Cascaded Residual Network (CARN) by using the cascade mechanism. Hui et al.~\cite{IMDN} proposed an Information Multi-Distillation Network (IMDN) by using the distillation and selective fusion strategy. MADNet~\cite{MADNet} used a dense lightweight network to enhance multi-scale feature representation and learning. Xiao et al.~\cite{LAINet} proposed a simple but effective deep lightweight model for SISR, which can adaptively generate convolutional kernels based on the local information of each position. However, the performance of these lightweight models is not ideal since they disable obtaining larger receptive fields and global information.

\subsection{Transformer-based SISR}
In order to model the long-term dependence of images, more and more researchers pay attention to Transformer, which was first used in the field of NLP. Recently, many Transformer-based methods have been proposed for computer vision tasks, which also promote the development of SISR. For example, Chen et al.~\cite{IPT} proposed a pre-trained Image Processing Transformer for image restoration. Liang et al.~\cite{r19} proposed a SwinIR by directly migrating the Swin Transformer to the image restoration task and achieved excellent results. Lu et al.~\cite{lu2021efficient} proposed an Effective Super-resolution Transformer (ESRT) for SISR, which reduces GPU memory consumption through a lightweight Transformer and feature separation strategy. However, all these models do not fully consider the fusion of CNN and Transformer, thus difficult to achieve the best balance between model size and performance.

\begin{figure*}[t]
	\centerline{\includegraphics[width=17.5cm, trim=0 20 0 0]{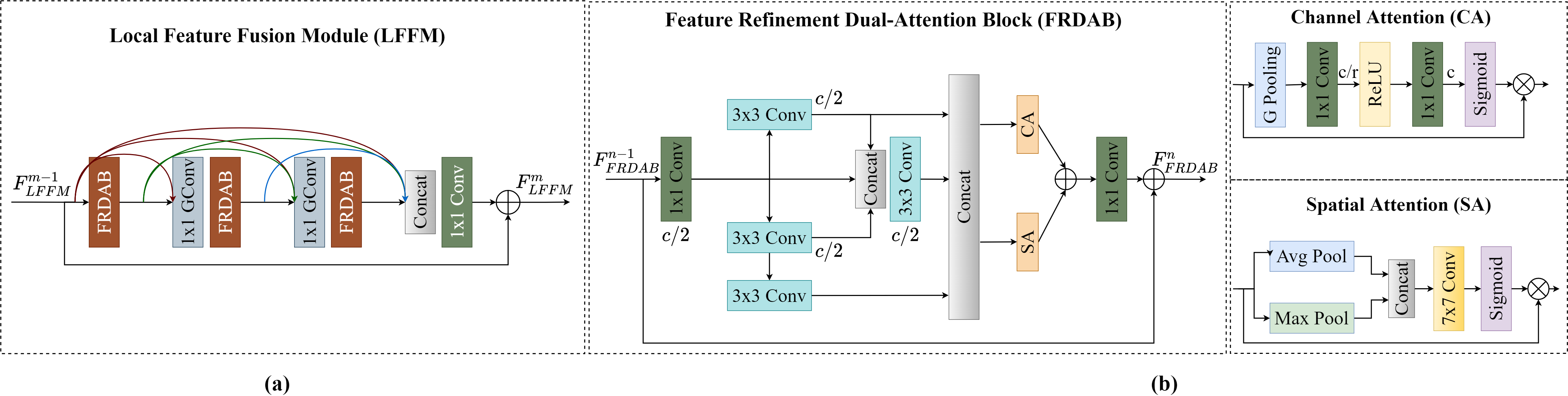}}
	\caption{The architecture of the proposed Local Feature Fusion Module (LFFM) and Feature Refinement Dual-Attention Block (FRDAB).}
	\vspace{-0.3cm}
	\label{Module}
\end{figure*}

\section{Lightweight Bimodal Network (LBNet)}
\subsection{Network Architecture}
As shown in Figure~\ref{LBNet}, Lightweight Bimodal Network (LBNet) is mainly composed of Symmetric CNN, Recursive Transformer, and reconstruction module. Specifically, Symmetric CNN is proposed for local feature extraction and Recursive Transformer is designed to learn the long-term dependence of images. We define $I_{LR}$, $I_{SR}$, and $I_{HR}$ as the input LR image, the reconstructed SR image, and the corresponding HR image, respectively. At the head of the model, a $3 \times 3$ convolutional layer is applied for shallow feature extraction
\begin{equation}\label{eq1}
\small
F_{sf} = f_{sf}(I_{LR}),
\end{equation}
where ${f_{sf}}(\cdot)$ represents the convolutional layer, ${F_{sf}}$ is the extracted shallow features. Then, the extracted shallow features will be sent to Symmetric CNN for local feature extraction
\begin{equation}\label{eq2}
\small
F_{CNN} = f_{CNN}(F_{sf}),
\end{equation}
where $f_{CNN}(\cdot)$ represents the Symmetric CNN and $F_{CNN}$ represents the extracted local features. Symmetric CNN is one of the most important components in LBNet, which consists of several pairs of parameter-sharing Local Feature Fusion Modules (LFFMs) and channel attention modules. All of these modules will be introduced in the next section.

After that, all these features will be sent to the Recursive Transformer for long-term dependence learning
\begin{equation}
\small
F_{RT} = f_{RT}(F_{CNN}),
\end{equation}
where $f_{RT}(\cdot)$ is the Recursive Transformer and $F_{RT}$ is the feature enhanced by global information. Finally, the refined features $F_{RT}$ and the shallow features $F_{sf}$ are added and sent to the reconstruction module for SR image reconstruction
\begin{equation}\label{eq3}
\small
I_{SR} = f_{build}(F_{sf} + F_{RT}),
\end{equation}
where $f_{build}(\cdot)$ is the reconstruction module, which composed of a $3 \times 3$ convolutional layer and a pixel-shuffle layer. 

During training, LBNet is optimized with $L1$ loss function. Given a training dataset $\left \{I_{LR}^{i}, I_{HR}^{i}  \right \}_{i=1}^{N}$, we solve
\begin{equation}
\small
\hat{\theta} = \arg\,\min\limits_{\theta}\, \frac{1}{N}\sum_{i=1}^{N}  \left \| F_{\theta}(I_{LR}^{i}) -  I_{HR}^{i} \right \|_{1},
\end{equation}
where $\theta$ denotes the parameter set of our proposed LBNet, $F(I_{LR}) = I_{SR}$ is the reconstruct SR image, and $N$ is the number of the training images.

\subsection{Symmetric CNN}
Symmetric CNN is specially designed for local feature extraction, which mainly consists of some paired parameter-sharing Local Feature Fusion Modules (LFFMs) and Channel Attention (CA) modules. The parameter-sharing of every two symmetrical modules can better balance the parameters and performance. In addition, each pair of parameter-sharing modules will be fused through the channel attention module, so that the extracted features can be fully utilized.

As shown in Figure~\ref{LBNet}, Symmetric CNN is a dual-branch network. The shallow feature ${F_{sf}}$ will first be sent to the top branch and the outputs of each LFFM in the top branch will serve as part of the input of the corresponding LFFM in the down branch. The complete operation can be defined as
\begin{equation}
\small
F_{LFFM}^{T,1} = f_{LFFM}^{T,1}({F_{sf}}),~i=1,
\vspace{-0.3cm}
\end{equation}
\begin{equation}
\small
F_{LFFM}^{T,i} = f_{LFFM}^{T,i}(F_{LFFM}^{T, i-1}),~i = 2,...,n,
\vspace{-0.3cm}
\end{equation}
\begin{equation}
\small
F_{LFFM}^{D,i} = f_{LFFM}^{D,i}(F_{LFFM}^{D, i-1} + f_{CA}^i(F_{LFFM}^{T,i})),~i = 2,...,n,
\end{equation}
where $f_{LFFM}^{T,i}(\cdot)$ and $f_{LFFM}^{D,i}(\cdot)$ represent the $i$-th LFFM in the top and down-branch, respectively. $f_{CA}^i(\cdot)$ denotes the $i$-th channel attention module. It is worth noting that when $i=1$, $F_{LFFM}^{D,1} = f_{LFFM}^{D,1}(F_{LFFM}^{T,n} + f_{CA}^{1}(F_{LFFM}^{T,1}))$. Moreover, the weight sharing strategy is applied on the paired modules, thus $f_{LFFM}^{D,i}(\cdot) = f_{LFFM}^{T,i}(\cdot)$. Finally, the outputs of all these LFFMs are concatenated and a $1 \times 1$ convolutional layer is used for feature fusion and compression. Therefore, the most effective features extracted at different levels will be sent to the next part to learn the long-term dependence of images.

\textbf{Local Feature Fusion Module (LFFM)}. LFFM is the core component of Symmetric CNN. As shown in Figure~\ref{Module} (a), LFFM is essentially an improved version of DenseBlock~\cite{DenseNet}. Different from DenseBlock, (1) we use FRDAB to replace the original convolutional layer to make it have a stronger feature extraction ability; (2) we introduce a 1 $\times$ 1 group convolutional layer before each FRDAB for dimensionality reduction; (3) local residual learning is introduced to further promote the transmission of information. The complete operation of LFFM can be defined as
\begin{equation}
\small
F_{FRB}^1 = f_{FRB}^1(F_{LFFM}^{m-1}),
\vspace{-0.3cm}
\end{equation}
\begin{equation}
\small
F_{FRB}^2 = f_{FRB}^2(f_{gc}^1([F_{LFFM}^{m-1},F_{FRB}^1])),
\vspace{-0.3cm}
\end{equation}
\begin{equation}
\small
F_{FRB}^3 = f_{FRB}^3(f_{gc}^2([F_{LFFM}^{m-1},F_{FRB}^1,F_{FRB}^2])),
\vspace{-0.3cm}
\end{equation}
\begin{equation}
\small
F_{LFFM}^m = f_{LFFM}^{m-1} + f_{1 \times 1}([F_{LFFM}^{m-1},F_{FRB}^1,F_{FRB}^2,F_{FRB}^3]),
\end{equation}
where $F_{FRB}^i$ represents the output of the $i$-th ($i=1,2,3$) FRDAB module in LFFM. $f_{gc}^j(\cdot)$ means the $j$-th ($j=1,2$) group convolutional layer followed by FRDAB. $F_{LFFM}^{m - 1}$ and $F_{LFFM}^{m}$ represent the input and output of the $m$-th LFFM module, respectively.

\textbf{Feature Refinement Dual-Attention Block (FRDAB)}. As shown in Figure~\ref{Module} (b), FRDAB is a dual-attention block, which specially designed for feature refinement. Specifically, the multi-branch structure is designed for feature extraction and utilization. In this part, the feature will be sent to two branches and each branch uses a different number of convolutional layers to change the size of the receptive field to obtain different scales features. c/2 indicates the operations of halving the outputs. After that, channel attention is used to extract channel statistics for re-weighting in the channel dimension and spatial attention is used to re-weighted the pixel according to the spatial context relationship of the feature map. Finally, the output of these two attention operations is fused by the addition operation. With the help of this method, the finally obtained features will show a stronger suppression of the smooth areas of the input image.

\subsection{Recursive Transformer}
\begin{figure}[t]
	\centerline{\includegraphics[width=6.5cm, trim=0 20 0 0]{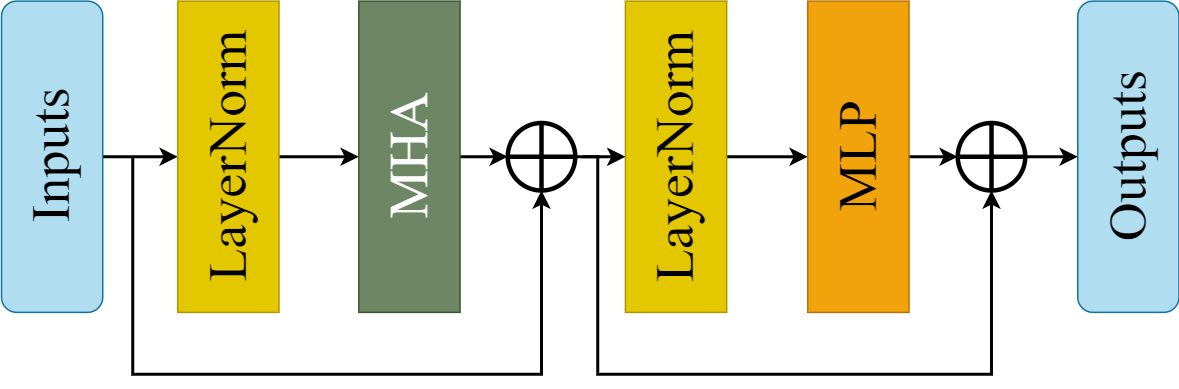}}
	\caption{The architecture of the Transformer Module (TM).}
	\vspace{-0.3cm}
	\label{TM}
\end{figure}

As we mentioned before, Symmetric CNN is designed for local feature extraction. However, this is far from enough to reconstruct high-quality images since the depth of the lightweight network makes it difficult to have a large enough receptive field to obtain global information. To solve this problem, we introduce Transformer to learn the long-term dependence of images and propose a Recursive Transformer (RT). Different from previous methods, we introduced the recursive mechanism to allow the Transformer to be fully trained without greatly increasing GPU memory consumption and model parameters. As shown in Figure~\ref{LBNet}, RT is located before the reconstruction module, which consists of two Transformer Modules (TM) and two convolutional layers. The completed operation of RT can be defined as
\begin{equation}
\small
F_{RT} = f_{3 \times 3}(f_{TM2}^ \circlearrowright (f_{3 \times 3}(f_{TM1}^ \circlearrowright (F_{CNN})))),
\end{equation}
where $f_{3 \times 3}(\cdot)$ and $f_{TM}(\cdot)$ represent the convolutional layer and the TM, respectively. $\circlearrowright$ denotes the recurrent connection, which means that the output of TM will be served as its new input and looped $S$ times. As for the TM, we only use the encoding part of the standard Transformer structure inspired by ESRT~\cite{lu2021efficient}. As shown in Figure~\ref{TM}, TM is mainly composed of two layer normalization layers, one Multi-Head Attention (MHA), and one Multi-Layer Perception (MLP). Define the input embeddings as $F_{in}$, the output embeddings $F_{out}$ can be obtained by
\begin{equation}
\small
F_{mid} = F_{in} + f_{MHA}(f_{norm}(F_{in})),
\vspace{-0.3cm}
\end{equation}
\begin{equation}
\small
F_{out} = F_1 + f_{MLP}(f_{norm}(F_{mid})),
\end{equation}
where $f_{norm}(\cdot)$ represents the layer normalization operation. $f_{MHA}(\cdot)$ and $f_{MLP}(\cdot)$ represent the MHA and MLP modules, respectively. Like ESRT, we project the input feature map of MHA into $Q$, $K$, and $V$ through a linear layer to reduce GPU memory consumption. Meanwhile, feature reduction strategy is also used to further reduce the memory consumption of the Transformer. Following~\cite{attention}, each head of the MHA must perform a scaled dot product attention, and then concatenate all the outputs and perform a linear transformation to obtain the output. Where the scaled dot product attention can be expressed as
\begin{equation}
\small
Attention(Q,K,V) = softmax (\frac{{Q{K^T}}}{{\sqrt {{d_k}} }})V.
\vspace{-0.1cm}
\end{equation}

This is the first attempt of the recursive mechanism in Transformer. With the help of this strategy, we can fully train and utilize Transformer without increasing the parameters and GPU memory consumption of the model. We will further discuss its effectiveness in the next section.

\section{Experiments}
\subsection{Datasets and Evaluation Metrics}
Following previous works, we use DIV2K as training set. For evaluation, we use five benchmark test datasets to validate the effectiveness of LBNet, including Set5~\cite{Set5}, Set14~\cite{Set14}, BSDS100~\cite{BSDS100}, Urban100~\cite{Urban100}, and Manga109~\cite{Manga109}. In addition, PSNR and SSIM are used as evaluation indicators to evaluate the performance of SR images on the Y channel of the YCbCr color space.

\subsection{Implementation Details}
To obtain LR images, we use bicubic interpolation to downsample the HR image and enhancing the training data with random rotation and horizontal flipping. During training, we randomly crop $48 \times 48$ patches from the training set as the inputs. The initial learning rate is $2 \times {10^{ - 4}}$ and finally dropped to $6.25 \times {10^{ - 6}}$ by cosine annealing. Meanwhile, the model is trained by Adam optimizer under the PyTorch framework with a NVIDIA RTX 2080Ti GPU. In the final model, the input and output channels of each part are set to $32$, three ($n=3$) LFFMs are used, and Transformer module recurses 2 times ($S=2$). Moreover, we also built a tiny version, named LBNet-T, which only consists of two ($n=2$) LFFMs. Meanwhile, the number of channels in LBNet-T has also been reduced to $18$.

\begin{table*}[t]
    \centering
    \begin{adjustbox}{width=1\linewidth}
    \begin{tabular}{|l|c|c|c|c|c|c|c|c|}
    \hline
    \multirow{2}{*}{Method} & \multicolumn{1}{l|}{\multirow{2}{*}{Scale}} & \multirow{2}{*}{Params}& \multirow{2}{*}{Mult-Adds} & Set5 & Set14 & BSD100 & Urban100 & Manga109 \\
    \cline{5-9} 
    & \multicolumn{1}{l|}{} & &  & PSNR/SSIM & PSNR/SSIM & PSNR/SSIM & PSNR/SSIM & PSNR/SSIM \\
    \hline
    \hline
    VDSR~\cite{VDSR}           	&\multirow{11}{*}{$\times 3$}							& 665K	&612.6G	& 33.66/0.9213        &29.77/0.8314   		   	   & 28.82/0.7976    & 27.14/0.8279			& 32.01/0.9310     \\
    DRCN~\cite{DRCN}      		&							& 1774K&17974.3G	& 33.82/0.9226        & 29.76/0.8311    		   & 28.80/0.7963    & 27.15/0.8276			& 32.31/0.9328    \\
    DRRN~\cite{DRRN}      		&							& 297K	&6796.9G	& 34.03/0.9244        & 29.96/0.8349    		   & 28.95/0.8004    & 27.53/0.8378			& 32.74/0.9390     \\		
    IDN~\cite{IDN}     			&							& 553K &56.3G		& 34.11/0.9253        & 29.99/0.8354    		   & 28.95/0.8013    & 27.42/0.8359			& 32.71/0.9381   \\
    CARN~\cite{CARN}      		& 							& 1592K&118.8G	& 34.29/0.9255        &30.29/0.8407   		   	   & 29.06/0.8034    & 28.06/0.8493			& 33.43/0.9427    \\
    IMDN~\cite{IMDN}      		& 							& 703K	&71.5G		& 34.36/0.9270       & 30.32/0.8417  		   	   & 29.09/0.8046    & 28.17/0.8519   			& 33.61/0.9445 \\
    AWSRN-M~\cite{AWSRN}      	& 							& 1143K	 &116.6G		& {\color{blue}34.42}/{\color{blue}0.9275}       & 30.32/{\color{blue}0.8419}  		   	   & {\color{red}29.13}/{\color{blue}0.8059}    & {\color{blue}28.26}/{\color{blue}0.8545}   		& 33.64/{\color{blue}0.9450} \\
    MADNet~\cite{MADNet}      	&							& 930K	&88.4G		& 34.16/0.9253        & 30.21/0.8398  		   	   & 28.98/0.8023    & 27.77/0.8439         		& -    \\
    GLADSR~\cite{GLADSR}      &							& 821K	 &88.2G	& 34.41/0.9272        & {\color{blue}30.37}/0.8418   		   	   & 29.08/0.8050    & 28.24/0.8537   		& - \\
    SMSR~\cite{SMSR}      		&							& 993K	 &156.8G& 34.40/0.9270       &30.33/0.8412  		   	   & 29.10/0.8050    & 28.25/0.8536   		& {\color{blue}33.68}/0.9445\\
    LAPAR-A~\cite{LAPAR}      	&							& 594K	&114.0G	& 34.36/0.9267        & 30.34/{\color{red}0.8421}   		   	   &{\color{blue}29.11}/0.8054    & 28.15/0.8523			& 33.51/0.9441   \\
    \textbf{LBNet-T (Ours)}  				&				&407K 	&22.0G		&34.33/0.9264 	&30.25/0.8402   	&29.05/0.8042   &28.06/0.8485    &33.48/0.9433  \\
    \textbf{LBNet (Ours)}  				&					&736K 	&68.4G		& {\color{red}34.47}/{\color{red}0.9277} 	&{\color{red}30.38}/0.8417   &{\color{red}29.13}/{\color{red}0.8061}    &{\color{red}28.42}/{\color{red}0.8559}    &{\color{red}33.82}/{\color{red}0.9460}  \\
    \hline
    \hline
    VDSR~\cite{VDSR}           	&\multirow{11}{*}{$\times 4$}							& 665K &612.6G	& 31.35/0.8838        & 28.01/0.7674     		   & 27.29/0.7251    & 25.18/0.7524			& 28.83/0.8809     \\
    DRCN~\cite{DRCN}      		&							& 1774K&17974.3G	& 31.53/0.8854        & 28.02/0.7670    		   & 27.23/0.7233    & 25.14/0.7510			& 28.98/0.8816     \\
    DRRN~\cite{DRRN}      		&							& 297K	  &6796.9G	& 31.68/0.8888        & 28.21/0.7720    		   & 27.38/0.7284    & 25.44/0.7638			&29.46/0.8960     \\		
    IDN~\cite{IDN}     			&							& 553K   &32.3G	& 31.82/0.8903        & 28.25/0.7730    		   & 27.41/0.7297    & 25.41/0.7632			&29.41/0.8942     \\
    CARN~\cite{CARN}      		& 							& 1592K  &	90.9G	& 32.13/0.8937        & 28.60/0.7806   		   	   & 27.58/0.7349    & 26.07/0.7837			& 30.42/0.9070    \\
    IMDN~\cite{IMDN}      		& 							& 715K	&40.9G		& {\color{blue}32.21}/0.8948       & 28.58/0.7811 		   	   & 27.56/0.7353    & 26.04/0.7838   			&30.45/0.9075 \\
    AWSRN-M~\cite{AWSRN}      	& 							& 1254K	 &72.0G		& {\color{blue}32.21}/{\color{blue}0.8954}       & {\color{blue}28.65}/{\color{red}0.7832}  		& 27.60/{\color{blue}0.7368}    & {\color{blue}26.15}/{\color{blue}0.7884}   		& {\color{blue}30.56}/{\color{blue}0.9093} \\
    MADNet~\cite{MADNet}      	&							& 1002K&54.1G	& 31.95/0.8917       & 28.44/0.7780   		   	   & 27.47/0.7327    & 25.76/0.7746			& -    \\
    GLADSR~\cite{GLADSR}      &							& 826K	 &52.6G	& 32.14/0.8940       & 28.62/0.7813   		   	   & 27.59/0.7361    & 26.12/0.7851  		& - \\
   SMSR~\cite{SMSR}      		&							& 1006K&89.1G& 32.12/0.8932 &28.55/0.7808  		   	   &27.55/0.7351    &26.11/0.7868   		& 30.54/0.9085\\
    LAPAR-A~\cite{LAPAR}      	&							& 659K	&94.0G		&32.15/0.8944        &28.61/{\color{blue}0.7818}   		   	   &{\color{blue}27.61}/0.7366    & 26.14/0.7871			&30.42/0.9074    \\
    \textbf{LBNet-T (Ours)}  			&					&410K 	&12.6G		&32.08/0.8933 &28.54/0.7802  	   	   &27.54/0.7358     &26.00/0.7819  &30.37/0.9059     \\
    \textbf{LBNet (Ours)}  				 &					&742K 	&38.9G		&{\color{red}32.29}/{\color{red}0.8960}  	 &{\color{red}28.68}/{\color{red}0.7832}  	  & {\color{red}27.62}/{\color{red}0.7382}    & {\color{red}26.27}/{\color{red}0.7906}    			&{\color{red}30.76}/{\color{red}0.9111}     \\
    \hline
    \end{tabular}
    \end{adjustbox}
    \vspace{-0.2cm}
    \caption{Average PSNR/SSIM comparison. The best and second best results are highlighted with \textcolor{red}{red} and \textcolor{blue}{blue}, respectively.}
    \vspace{-0.15cm}
    \label{SOTA}
\end{table*}

\subsection{Comparison with Lightweight SISR Models}
\begin{figure*}[t]
	\newlength\fsdttwofig
	\setlength{\fsdttwofig}{-3mm}
	\scriptsize
	\centering
	\begin{tabular}{cc}
		\begin{adjustbox}{valign=t}
		\tiny
			\begin{tabular}{c}
				\includegraphics[height=27mm,width=40mm]{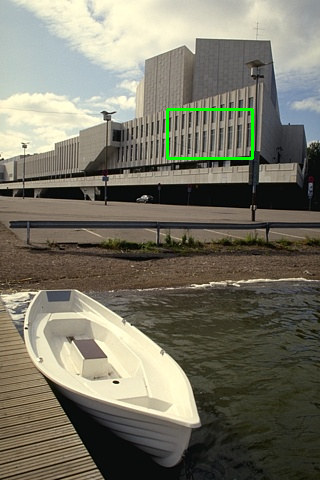}
				\\
				 BSDS100 ($4\times$):
				\\
				78004
				
			\end{tabular}
		\end{adjustbox}
		\hspace{-2.3mm}
		\begin{adjustbox}{valign=t}
		\tiny
			\begin{tabular}{cccccc}
				\includegraphics[height=10mm,width=24mm]{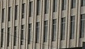} \hspace{\fsdttwofig} &
				\includegraphics[height=10mm,width=24mm]{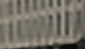} \hspace{\fsdttwofig} &
				\includegraphics[height=10mm,width=24mm]{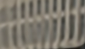} \hspace{\fsdttwofig} &
				\includegraphics[height=10mm,width=24mm]{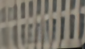} \hspace{\fsdttwofig} &
				\includegraphics[height=10mm,width=24mm]{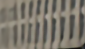} 
				\\
				HR \hspace{\fsdttwofig} &
				SRCNN \hspace{\fsdttwofig} &
				DRCN \hspace{\fsdttwofig} &
				IDN \hspace{\fsdttwofig} &
				CARN-M
				\\
				PSNR/SSIM \hspace{\fsdttwofig} &
				25.20/0.6840 \hspace{\fsdttwofig} &
				25.93/0.7266 \hspace{\fsdttwofig} &
				26.65/0.7532 \hspace{\fsdttwofig} &
				26.46/0.7471
				\\
				\includegraphics[height=10mm,width=24mm]{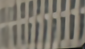} \hspace{\fsdttwofig} &
				\includegraphics[height=10mm,width=24mm]{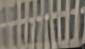} \hspace{\fsdttwofig} &
				\includegraphics[height=10mm,width=24mm]{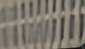} \hspace{\fsdttwofig} &
				\includegraphics[height=10mm,width=24mm]{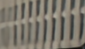} \hspace{\fsdttwofig} &
				\includegraphics[height=10mm,width=24mm]{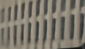}  
				\\ 
				CARN \hspace{\fsdttwofig} &
				IMDN \hspace{\fsdttwofig} &
				MADNet \hspace{\fsdttwofig} &
				LBNet-T (Ours)  \hspace{\fsdttwofig} &
				LBNet (Ours)  
				\\
				26.72/0.7572 \hspace{\fsdttwofig} &
				26.43/0.7509 \hspace{\fsdttwofig} &
				26.40/0.7451 \hspace{\fsdttwofig} &
				26.88/0.7602 \hspace{\fsdttwofig} &
				\textbf{27.00}/\textbf{0.7649} \hspace{\fsdttwofig} 
				\\
			\end{tabular}
		\end{adjustbox}
		\vspace{0.5mm}
		\\
		\begin{adjustbox}{valign=t}
		\tiny
			\begin{tabular}{c}
				\includegraphics[height=27mm,width=40mm]{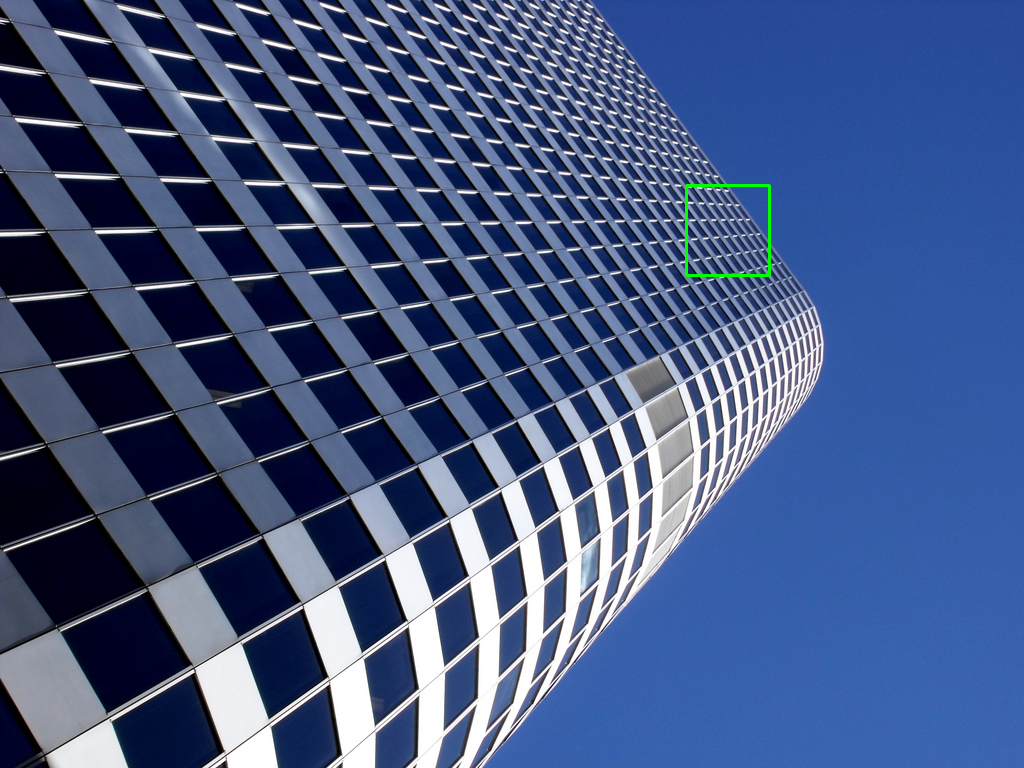}
				\\
				Urban100 ($4\times$):
				\\
				img\_005
			\end{tabular}
		\end{adjustbox}
		\hspace{-2.3mm}
		\begin{adjustbox}{valign=t}
		\tiny
			\begin{tabular}{cccccc}
				\includegraphics[height=10mm,width=24mm]{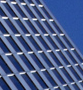} \hspace{\fsdttwofig} &
				\includegraphics[height=10mm,width=24mm]{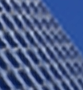} \hspace{\fsdttwofig} &
				\includegraphics[height=10mm,width=24mm]{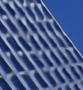} \hspace{\fsdttwofig} &
				\includegraphics[height=10mm,width=24mm]{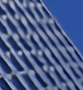} \hspace{\fsdttwofig} &
				\includegraphics[height=10mm,width=24mm]{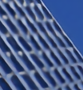} 
				\\
				HR \hspace{\fsdttwofig} &
				SRCNN \hspace{\fsdttwofig} &
				DRCN \hspace{\fsdttwofig} &
				IDN \hspace{\fsdttwofig} &
				CARN-M
				\\
				PSNR/SSIM \hspace{\fsdttwofig} &
				25.12/0.8860 \hspace{\fsdttwofig} &
				26.79/0.9328 \hspace{\fsdttwofig} &
				27.64/0.9466 \hspace{\fsdttwofig} &
				26.96/0.9409 
				\\
				\includegraphics[height=11mm,width=24mm]{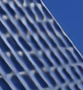} \hspace{\fsdttwofig} &
				\includegraphics[height=11mm,width=24mm]{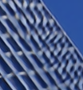} \hspace{\fsdttwofig} &
				\includegraphics[height=11mm,width=24mm]{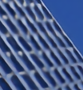} \hspace{\fsdttwofig} &
				\includegraphics[height=11mm,width=24mm]{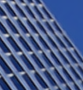} \hspace{\fsdttwofig} &
				\includegraphics[height=11mm,width=24mm]{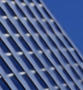}  
				\\ 
				CARN \hspace{\fsdttwofig} &
				IMDN \hspace{\fsdttwofig} &
				MADNet \hspace{\fsdttwofig} &
				LBNet-T (Ours)  \hspace{\fsdttwofig} &
				LBNet (Ours) 
				\\
				27.70/0.9500 \hspace{\fsdttwofig} &
				27.35/0.9472 \hspace{\fsdttwofig} &
				27.09/0.9421 \hspace{\fsdttwofig} &
				28.57/0.9539 \hspace{\fsdttwofig} &
				\textbf{28.70}/\textbf{0.9581} \hspace{\fsdttwofig} 
				\\
			\end{tabular}
		\end{adjustbox}
	\end{tabular}
	\vspace{-3mm}
	\caption{Visual comparison with other SISR models. Obviously, our LBNet can reconstruct realistic SR images with accurate edges.}
	\vspace{-3mm}
\label{Visual_comparison}
\end{figure*}

In Table~\ref{SOTA}, we compare our LBNet with 11 advanced lightweight SISR models. Most of them achieve the best results at the time in the lightweight SISR task. According to the table, we can clearly observe that our LBNet achieves the best results, and our tiny version LBNet-T also achieves the best results under the same parameter level. In addition, the number of parameters and Mult-Adds of LBNet and LBNet-T are also very low, which proves the efficiency of our models. Meanwhile, we also provide the visual comparison between LBNet and other lightweight SISR models in Figure~\ref{Visual_comparison}. Obviously, SR images reconstructed by our LBNet have richer detailed textures with better visual effects. This further validates the effectiveness of our proposed LBNet. 

\begin{figure}[t]
	\centerline{\includegraphics[width=8.8cm, trim=0 20 20 40]{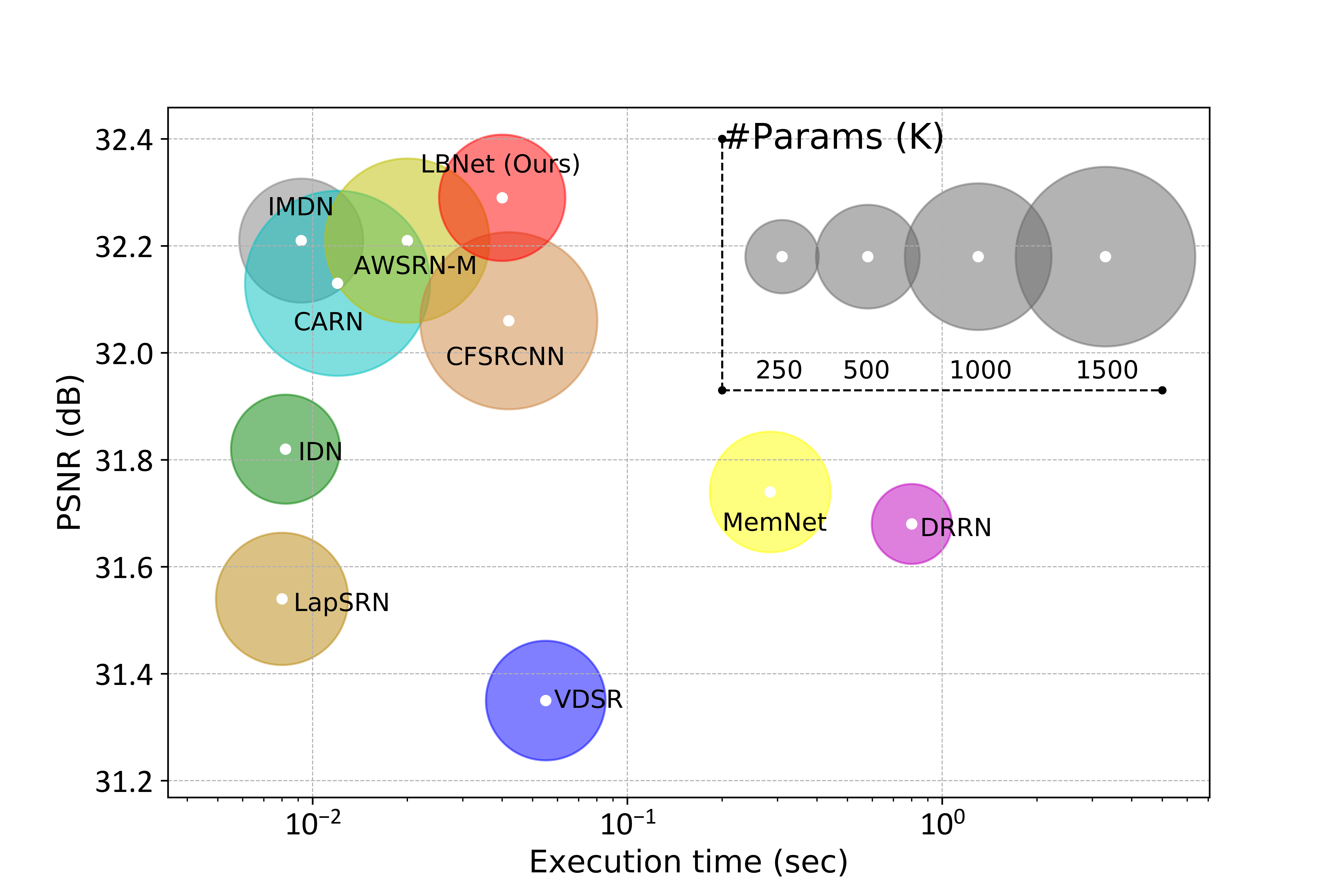}}
	\caption{Model complexity study on Set5 ($\times 4$).}
	\vspace{-0.4cm}
	\label{ExecTime}
\end{figure}

\subsection{Model Complexity Studies}
As can be seen from Table~\ref{SOTA}, our model achieves an excellent balance between model size, performance, and Mult-Adds. In addition, the execution time of the model is also an important indicator to measure the complexity of the model. To make a more intuitive comparison with other models, we provide the trade-offs between model performance, parameter quantity, and execution time in Figure~\ref{ExecTime}. Obviously, our LBNet achieved the best PSNR results under the premise of comparable execution time and parameters. This further illustrates that LBNet is an efficient and lightweight SISR model.

\subsection{Ablation Study}
\textbf{Symmetric CNN Investigations.} In Symmetric CNN, each pair of LFFMs will be fused through a channel attention module. In order to verify the effectiveness of this fusion method, we conduct a series of experiments to study the different feature interaction methods in Symmetric CNN. It is worth noting that we removed the Transformer part of the model to speed up the training time in this ablation study. Table~\ref{Ablation1} shows the results of these different feature interaction methods. Among them, FF means the ordinary concatenate operation without attention module, SA and CA indicate the use of spatial attention and channel attention for fusion, respectively. Obviously, our method achieves the best results with the same parameters and Multi-Adds level. This fully demonstrates the effectiveness of our method. 

\begin{table}[t]
	\centering
		\begin{adjustbox}{width=1\linewidth} 
		\begin{tabular}{@{}ccccccc@{}}
			\toprule
			Scale   			&FF		 &SA	              & CA
			&Params     	    &Mult-Adds& PSNR/SSIM	      	\\ 
			\midrule
			$\times 4$           &\Checkmark  	&\XSolid    		&\XSolid		&96.1K	       		&10.01G		    &25.28/0.7601          \\
			$\times 4$      	&\XSolid  	&\Checkmark    	&\XSolid	 	  &96.3K		        &10.03G        &25.31/0.7614                \\
			$\times 4$     	&\XSolid  	&\XSolid   		&\Checkmark 	
			&96.5K				&10.01G		&\textbf{25.36}/\textbf{0.7622}   \\ 
			\bottomrule 
		\end{tabular}
		\end{adjustbox}
		\vspace{-0.2cm}
		\caption{Study of the different feature interaction schemes in Symmetric CNN on Urban100 ($\times 4$). Best results are highlighted.}
		\vspace{-0.3mm}
		\label{Ablation1}
\end{table}

\begin{table}[t]
	\centering
	    \begin{adjustbox}{width=1\linewidth} 
		\begin{tabular}{@{}l|ccc@{}}
			\toprule
			Method   				&Params &Mult-Adds 	&PSNR/SSIM	      	\\ 
			\midrule
			LBNet+RCAB   	 	&228K	&23.7G			&29.94/0.9002   \\ 
			LBNet+IMDB      	&295K	 &31.3G 		&30.21/0.9043         \\
			LBNet+FRDAB (Ours)   	&365K	&38.9G			&\textbf{30.33}/\textbf{0.9059}   \\
			\bottomrule 
		\end{tabular}
		\end{adjustbox}
		\vspace{-0.2cm}
		\caption{Performance comparisons of FRDAB and other basic units on Manga109 for $\times 4$ SR. The best results are highlighted.}
		\vspace{-0.3cm}
		\label{Ablation2}
\end{table}

\begin{table*}
\centering
\begin{adjustbox}{width=1\linewidth} 
\begin{tabular}{l|cccccccc}
\toprule
Method & Params & Multi-Adds & Set5  & Set14 & BSD100 & Urban100 & Manga109 & Average \\
\midrule
SwinIR & 897K   & 49.6G      & 32.44/0.8976 & 28.77/0.7858 & 27.69/0.7406  & 26.47/0.7980    & 30.92/0.9151    & 29.26/0.8274   \\ 
ESRT   & 751K   & 67.7G      & 32.19/0.8947 & 28.69/0.7833 & 27.69/0.7379  & 26.39/0.7962    & 30.75/0.9100    & 29.14/0.8244   \\ 
LBNet (Ours)  & \textbf{742K}   & \textbf{38.9G}      & 32.29/0.8960 & 28.68/0.7832 & 27.62/0.7382  & 26.27/0.7906    & 30.76/0.9111    & 29.12/0.8238   \\ 
\bottomrule 
\end{tabular}
\end{adjustbox}
\vspace{-0.2cm}
\caption{Comparison with other Transformer-based methods. LBNet can achieve competitive results with fewer parameters and Multi-Adds.}
\vspace{-0.3cm}
\label{Trans}
\end{table*}

To verify the effectiveness of our FRDAB, we replace FRDAB with some commonly used feature extraction modules in lightweight SISR models, like IMDB~\cite{IMDN} and RCAB~\cite{RCAN}. It can be seen from Table~\ref{Ablation2} that although FRDAB will bring a little increase of parameters and Mult-Adds, its performance has been significantly improved. This gain is considerable, which benefit for lightweight models construction. This fully proves that FRDAB is an effective feature extraction module.

\begin{table}[t]
	\centering
		\begin{adjustbox}{width=1\linewidth} 
		\begin{tabular}{@{}r|cccc@{}}
			\toprule
			Method   			&Params &Mult-Adds 	&Running time	&PSNR/SSIM	      	\\ 
			\midrule
			w/o RT      			&365K	 &38.9028G 	&0.0168s		&32.07/0.8929          \\
			with RT     	 			&742K	&38.9032G	&0.0274s				&\textbf{32.23}/\textbf{0.8949}   \\ 
			\bottomrule 
		\end{tabular}
		\end{adjustbox}
		\vspace{-0.2cm}
		\caption{Study of Recursive Transformer (RT) on Set5 dataset ($\times 4$).}
		\vspace{-0.3cm}
		\label{Ablation3}
\end{table}

\begin{table}[t]
	\centering
		\begin{adjustbox}{width=1\linewidth}  
		\begin{tabular}{@{}c|cccc@{}}
			\toprule
			Method   			&Params &Mult-Adds 	&Running time	&PSNR/SSIM	      	\\ 
			\midrule
			TM-0      			&741.7K	 &38.9032G 	    &0.0274s				&32.23/0.8949          \\
			TM-1      	 		&741.7K	&38.9036G		&0.0356s				&32.27/0.8958   \\
			\textbf{TM-2}      	 		& \textbf{741.7K}	&  \textbf{38.9039G}		& \textbf{0.0401s}				& \textbf{32.29/0.8960}   \\ 
			TM-3      	 		&741.7K	&38.9043G		&0.0516s				&32.30/0.8960   \\  
			\bottomrule 
		\end{tabular}
		\end{adjustbox}
		\vspace{-0.2cm}
		\caption{Study on the recursion times of Transformer Module (TM) on Set5 ($\times 4$). The final version is highlighted.}
		\vspace{-0.3cm}
		\label{Ablation4}
\end{table}

\paragraph{Recursive Transformer Investigations.} In order to learn the long-term dependence of images, we introduced the Transformer and proposed a Recursive Transformer (RT). To verify the effectiveness of the proposed RT, we remove RT and provide the results in Table~\ref{Ablation3}. According to the table, we can clearly observe that the introduced RT will increase the number of parameters. However, the increase in Multi-add and execution time is insignificant. In addition, the experiment also shows that the PSNR/SSIM results of the model with RT have been significantly improved. It shows that the introduced Transformers can improve the learning capabilities of the model thus improving the model performance.

Different from previous Transformers, we introduced the recursive mechanism in the Transformer Module (TM). This design can make the Transformer more fully utilized without increasing the number of model parameters. To verify the effectiveness of the recursive mechanism in Transformer, we conduct a series of experiments with different recursion times. In Table~\ref{Ablation4}, TM-N denotes TM recursed N times and TM-0 represents the flat model that no recursive mechanism is used. Obviously, as the number of recursion times increases, the performance of the model will further improve. Meanwhile, we also notice that when the number of recursion times is 3 (TM-3), the improvement of model performance is not obvious. Therefore, we set the recursion times as 2 ($S=2$) in the final model to achieve a better balance between mode performance, Multi-Adds, and execution time.

\paragraph{Comparison with Other Transformer-based Methods.} Recently, some Transformer-based methods have been proposed for SISR. In Table~\ref{Trans}, we provide a detailed comparison with SwinIR~\cite{r19} and ESRT~\cite{lu2021efficient}. According to the results, we can clearly observe that our LBNet achieved competitive results with fewer parameters and Multi-Adds. Although the PSNR result of LBNet is 0.14dB worse than SwinIR, it is worth noting that SwinIR uses an additional dataset (Flickr2K) for training. This is one of the key factors to further improve model performance. All these results further verify the effectiveness of LBNet.

\section{Conclusions}
In this paper, we proposed a Lightweight Bimodal Network  (LBNet) for SISR via Symmetric CNN and Recursive Transformer. Specifically, we proposed an effective Symmetric CNN for local feature extraction and proposed a Recursive Transformer to learn the long-term dependence of images. In Symmetric CNN, Local Feature Fusion Module (LFFM) and Feature Refinement Dual-Attention Block (FRDAB) are designed to ensure sufficient feature extraction and utilization. In Recursive Transformer, the recursive mechanism is introduced to fully train the Transformer, so the global information learned by the Transformer can further refine the features. In summary, LBNet elegantly integrates CNN and Transformer, achieving a better balance between the performance, size, execution time, and GPU memory consumption of the model.

\section*{Acknowledgments}
This work was supported in part by the National Key R\&D Program of China (No.2021YFE0203700), the National Natural Science Foundation of China (Nos.61972212, 61772568, and 62076139 and 61833011), the Natural Science Foundation of Jiangsu Province (No.BK20190089), the Six Talent Peaks Project in Jiangsu Province (No.RJFW-011), and the CRF (No.8730063).

\bibliographystyle{named}
\bibliography{ijcai22}

\end{document}